%% file: main.tex
\documentclass[acmtog,screen]{acmart}

\AtBeginDocument{%
  }

\input{preamble}

\copyrightyear{2026}
\acmYear{2026}
\setcopyright{cc}
\setcctype{by}
\acmConference[SA Conference Papers '26]{SIGGRAPH Asia 2026 Conference Papers}{December 01--04, 2026}{Kuala Lumpur, Malaysia}
\acmBooktitle{SIGGRAPH Asia 2026 Conference Papers (SA Conference Papers '26), December 01--04, 2026, Kuala Lumpur, Malaysia}
\acmDOI{10.1145/3829340.3842275}
\acmISBN{979-8-4007-2842-6/2026/12}

\acmSubmissionID{1837}

\usepackage{enumitem}

\hypersetup{
    colorlinks=true,      
    linkcolor=red,         
    citecolor=blue,        
    urlcolor=blue         
}

\begin{document}

\title{OmniFabric: Coherent UV Space Texture Synthesis for 3D Garment Reconstruction}


\author{Ding-Jiun Huang}
\affiliation{%
  \institution{Carnegie Mellon University}
  \country{United States of America}
}
\email{djhuang322@gmail.com}

\author{Yuanhao Wang}
\affiliation{%
  \institution{University of Washington}
  \country{United States of America}
}
\email{yuanhao4@cs.washington.edu}

\author{Cheng Zhang}
\affiliation{%
  \institution{Texas A\&M University}
  \country{United States of America}
}
\email{chzhang@tamu.edu}

\author{Hugo Bertiche}
\affiliation{%
  \institution{Google}
  \country{United States of America}
}
\email{hbertiche@google.com}

\author{Alexandru-Eugen Ichim}
\affiliation{%
  \institution{Google}
  \country{Switzerland}
}
\email{alexichim@google.com}

\author{Thabo Beeler}
\affiliation{%
  \institution{Google}
  \country{Switzerland}
}
\email{tbeeler@google.com}

\author{Fernando De la Torre}
\affiliation{%
  \institution{Carnegie Mellon University}
  \country{United States of America}
}
\email{ftorre@andrew.cmu.edu}

\renewcommand{\shortauthors}{Huang, et al.}

\input{sec/0_abstract}




\maketitle

\input{sec/1_intro}
\input{sec/2_relate}
\input{sec/4_method}
\input{sec/5_experiments}
\input{sec/6_conclusion}

\clearpage % 1. Flushes any pending floats and forces a new page

\input{sec/additional_qual}
\clearpage % 2. Forces the bibliography to wait until a completely fresh page

\bibliographystyle{ACM-Reference-Format}
\bibliography{main}

\appendix
\renewcommand{\appendixname}{Supplementary Material~\Alph{section}}
\setcounter{footnote}{0}
\setcounter{table}{0}
\setcounter{figure}{0}
\setcounter{section}{0}
\renewcommand\thesection{\Alph{section}}
\renewcommand{\thetable}{S\arabic{table}}  
\renewcommand{\thefigure}{S\arabic{figure}}

\vspace{5mm}
\centerline{\textbf{\LARGE{ {Supplementary Material}}}}

\input{sec/x_supp}

\end{document}

%% file: preamble.tex
\usepackage{xspace}
\usepackage{pifont} % http://ctan.org/pkg/pifont
\usepackage{multirow}

\usepackage{tocloft}

%% file: sec/0_abstract.tex
\begin{teaserfigure}
    \centerline{\includegraphics[width=1\linewidth]{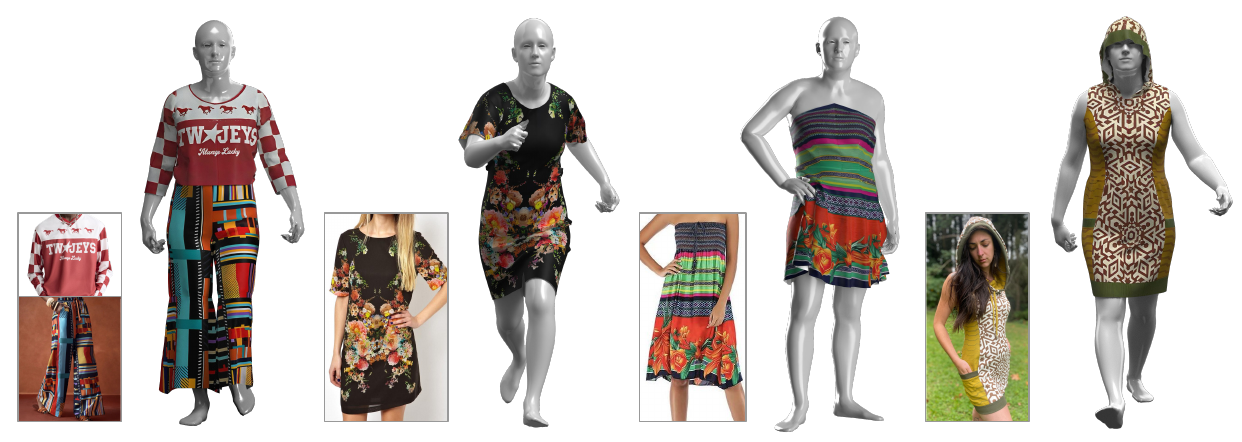}}
    \vspace{-1mm}
    \caption{\small Image-based 3D garment texture synthesis. Given a single in-the-wild clothing image, OmniFabric synthesizes high-quality, coherent fabric textures directly on garment sewing patterns. OmniFabric supports a wide range of textures and patterns, preserving fine details while maintaining global alignment with the input image. The resulting textured sewing patterns can be seamlessly simulated into 3D garments.
    }
    \Description{Four example pairs showing input images and simulation results.}
    \label{fig:teaser} 
    \vspace{2mm}
\end{teaserfigure}

\begin{abstract}
Automated generation of production-ready 3D garment assets from a single image is a central challenge in digital content creation. While recent generative models have significantly advanced 3D geometry reconstruction, synthesizing high-quality textures remains a bottleneck. Existing methods often bake environmental illumination and shadows directly into the texture map, or they fail to maintain global structural coherence, making the resulting assets unusable for physical simulation and relighting. In this work, we introduce OmniFabric, a novel approach that synthesizes globally coherent texture maps directly within the 2D sewing pattern space. Given a single reference image, our pipeline utilizes an estimated 3D mesh and generative priors of powerful Vision-Language Models (VLM) to establish a complete but coarse texture initialization across the unwrapped sewing patterns. We then leverage a specialized diffusion transformer, trained via an automated synthetic data engine and conditioned on 3D positional features, to refine this initialization directly in the canonical UV domain. This effectively removes distortion and baked-in artifacts to extract a clean and normalized texture map that preserves the original garment design. Extensive experiments demonstrate that OmniFabric significantly outperforms state-of-the-art baselines, yielding photorealistic 3D garments with high-quality textures. Project page: \href{https://humansensinglab.github.io/OmniFabric}{\textcolor{ACMDarkBlue}{https://humansensinglab.github.io/OmniFabric}}

\end{abstract}

%% file: sec/1_intro.tex
\section{Introduction}

Creating high-fidelity, simulation-ready 3D garments is essential for modern digital production, with applications spanning gaming~\cite{habermann2021real}, e-commerce~\cite{hwangbo2020effects,kim2013consumer}, and cinematic visual effects~\cite{hughes2007physical}. While traditional asset creation relies on labor-intensive manual modeling, there is growing interest in automating this process to synthesize assets directly from images. Despite recent strides in reconstructing accurate 3D garment geometry~\cite{bian2025chatgarment,nakayama2025aipparel,sarafianos2025garment3dgen,he2024dresscode,li2025dress,rong2025gaussian,wang2025garmentcrafter}, generating production-ready textures from a single observation remains an open problem.

Garment texture synthesis is uniquely difficult due to the complex dynamics of fabric, which frequently produce deep folds, self-occlusions, and extreme pose variations (see Figure~\ref{fig:teaser}). To be truly usable in downstream applications, a texture must not only look realistic in a static view but also be free from baked-in lighting and geometric distortion, capturing the garment's intrinsic albedo. However, current state-of-the-art 3D texturing frameworks~\cite{richardson2023texture,chen2023text2tex,zeng2024paint3d,zhao2025hunyuan3d,bensadoun2024meta} typically rely on multi-view diffusion priors to iteratively ``paint'' assets via direct projection onto draped 3D geometry. Consequently, unwrapping these folded surfaces inevitably causes large geometric distortion and occlusion. Furthermore, these models fail to disentangle intrinsic material properties from environmental illumination, baking transient lighting artifacts, such as shadows and geometry-induced wrinkles, directly into the texture map.

To circumvent these issues, recent work such as FabricDiffusion~\cite{zhang2024fabricdiffusion} has explored texture synthesis directly within the regularized, flat 2D sewing pattern domain, successfully yielding simulation-ready texture maps.
However, FabricDiffusion focuses exclusively on local texture synthesis by extracting tileable material patches from the source image, ignoring the global texture layout. Consequently, it struggles at reconstructing asymmetric designs, large-scale prints, or irregular spatial patterns. We argue that synthesizing a globally coherent texture map that faithfully reflects the visual identity of the input image requires leveraging the complete context of the reference observation, rather than relying on isolated patches. This presents a fundamental challenge: 
\begin{center}
\emph{How do we establish pixel-level correspondences between \\ a reference image and 2D sewing panels while transferring textures \\ that are free from baked-in artifacts and geometric distortions?}
\end{center}

In this paper, we propose \textbf{OmniFabric} to bridge this gap. We first employ a powerful image generation model~\cite{nanobananapro} to repose the input clothing into a canonical A-pose, aligning its silhouette with the frontal render of the garment mesh simulated from the predicted sewing patterns. We further harness the temporal and spatial consistency of large-scale video generation models~\cite{googledeepmind2025veo3web} to hallucinate coherent, multi-view observations of this A-pose garment. Projecting these globally consistent multi-view images onto the unwrapped sewing patterns establishes a complete, albeit coarse, texture initialization that preserves the garment's global design. Because this initial projection inevitably bakes visual artifacts such as geometric distortions and physics-induced wrinkles into the texture map, we introduce a subsequent refinement stage operating entirely within the sewing pattern space. We design an automatic data engine that curates an extensive and highly diverse synthetic dataset of textured sewing patterns, paired with simulated texture initializations containing baked-in artifacts and distortions. With the curated dataset, we train a specialized Diffusion Transformer (DiT) that acts as a texture normalizer in the 2D sewing pattern space. This model rectifies distortions, and strips away transient illumination to produce a clean and normalized texture map. Finally, we conduct comprehensive experiments to demonstrate the superiority of OmniFabric over existing 3D texturing baselines. In summary, the key contributions of our work are:
% \yuanhao{Need to think about contributions}
\begin{itemize}[itemsep=2pt,topsep=3pt,leftmargin=15pt]
    \item OmniFabric, a novel framework for synthesizing high-fidelity, globally coherent textures from a single in-the-wild image for 3D garment reconstruction.
    \item An automated data engine for creating realistic, diverse textured sewing patterns at scale, providing training data for learning complex garment textures.
    \item A strategy of using a canonical 3D mesh as spatial anchor to map image pixels to sewing pattern UV space, enabling globally aligned texture synthesis.
    \item A coarse-to-fine texture generation pipeline that initializes globally aligned textures, and further rectifies distortion and baked-in artifacts with a DiT-based UV space normalization network.
\end{itemize}
\label{sec:intro}

\paragraph{Remark.}
Generating 3D garment textures is not a new topic. While many prior works mainly focus on directly texturing 3D garment meshes, they often do not scale well to large collections of in-the-wild images and remain weakly connected to real-world garment production pipelines. We argue that fabric texture synthesis should be grounded in the UV space of sewing patterns. This enables the model to maintain strong global coherence while remaining closely aligned with how garments are constructed in practice.

%% file: sec/2_relate.tex
\section{Related Work}

\subsection{3D Garment Reconstruction}
Early approaches for garment reconstruction primarily relied on implicit functions~\cite{saito2019pifu,saito2020pifuhd,xiu2022icon} to approximate surfaces, yet they often struggle with complex topologies and loose-fitting clothing. More recent works leverage advanced representations like 3DGS~\cite{rong2025gaussian} and image-to-3D diffusion priors~\cite{wang2025garmentcrafter,luo2024garverselod} to achieve impressive visual fidelity. However, these methods typically produce rigid meshes or unstructured point clouds that are inherently unsuitable for physical cloth simulation. While methods like Garment3DGen~\cite{sarafianos2025garment3dgen} bridge this gap by utilizing template deformation to produce simulation-ready assets, they rely on a slow, computationally intensive per-asset optimization process.
To obtain simulation-ready assets, a significant line of work attempts to infer 2D sewing patterns directly from 3D point clouds or images. NeuralTailor~\cite{korosteleva2022neuraltailor} focused on reconstructing pattern structures from 3D geometry, and subsequent works~\cite{liu2023towards,chen2024panelformer} predict these patterns directly from single-view images using discriminative transformer architectures. To facilitate more structured generation, GarmentCode~\cite{korosteleva2023garmentcode} introduced a programmatic domain-specific language (DSL) for sewing patterns, enabling parametric control. Building on this representation, subsequent research has increasingly focused on advanced generative modeling. DressCode~\cite{he2024dresscode} synthesizes novel patterns from text prompts using a GPT-based framework, while AIpparel~\cite{nakayama2025aipparel} scales into a multimodal foundation model that handles complex pattern generation natively from mixed inputs. To further improve physical realism, Dress-1-to-3~\cite{li2025dress} incorporates differentiable physics simulators to optimize geometric alignment of the inferred patterns. Concurrently, other state-of-the-art frameworks directly predict these programmatic structures from images leveraging large generative models~\cite{bian2025chatgarment,zhou2025design2garmentcode,li2025garmentdiffusion}. While these methods have established sewing patterns as a robust domain for 3D animation, the synthesis of normalized, artifact-free textures for these panels remains an open challenge.  

\subsection{Generative 3D Texturing}
The rise of large-scale vision-language models~\cite{radford2021learning,rombach2022high,saharia2022photorealistic} has revolutionized 3D texture synthesis. Early optimization-based approaches either leverage CLIP \cite{radford2021learning} for optimizing texture maps of 3D models \cite{hong2022avatarclip,chen2022tango,michel2022text2mesh,mohammad2022clip}, or employ Score Distillation Sampling (SDS) to optimize texture maps by distilling gradients from a frozen 2D diffusion model~\cite{poole2022dreamfusion,metzer2023latent,chen2023fantasia3d}. While capable of generating coherent global structures, these methods tend to hallucinate generic textures without high-frequency details. To improve fidelity, recent research has shifted toward projection-based inpainting, by utilizing depth-conditioned Stable Diffusion to progressively paint 3D meshes from multiple viewpoints~\cite{richardson2023texture,chen2023text2tex,cao2023texfusion}. These frameworks iteratively project the current texture into screen space, inpaint missing regions using 2D priors, and project the result back. More recent systems like Paint3D~\cite{zeng2024paint3d}, MVPaint~\cite{cheng2025mvpaint} and Meta 3D TextureGen~\cite{bensadoun2024meta} further refine this process by synchronizing multi-view generation or employing coarse-to-fine UV refinement to minimize seams. Despite their popularity, a fundamental limitation persists across these general-purpose methods: they rely on priors trained on natural photography. Consequently, they inherently entangle illumination with surface appearance, inevitably ``baking in'' transient lighting effects—such as cast shadows and specular highlights—directly into the UV map. In addition, the generated textures are subject to distortion and blurring due to the complex surface topology, leading to suboptimal results. 

\begin{figure}[!t]
  \centering
  \includegraphics[width=0.94\columnwidth]{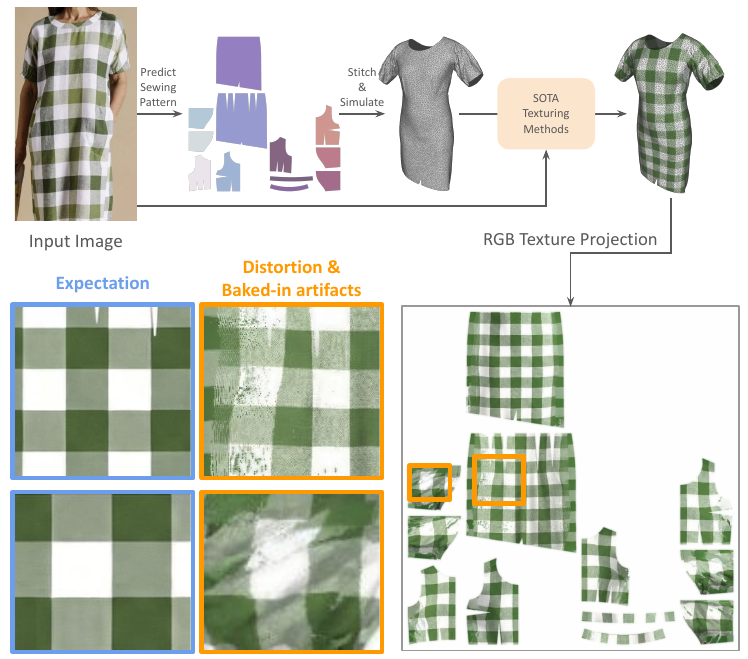}
  \vspace{-3mm}
  \caption{\textbf{Limits of baseline texturing methods.} While existing methods can synthesize textures for 3D assets with decent visual quality, they fail to create garment textures free from distortions and baked-in illumination artifacts, rendering the asset unsuitable for downstream applications. 
    }
    \Description{A figure showing the issues of existing 3D texturing methods.}
  \label{fig:motivation}
  \vspace{-2mm}
\end{figure}

In the specific area of garment texturing, FabricDiffusion~\cite{zhang2024fabricdiffusion} addresses some of these pitfalls by treating texture synthesis as a material extraction task. It utilizes a diffusion model to extract tileable, distortion-free material patches from a single image without lighting artifacts. However, because it relies on local texture tiling, FabricDiffusion fails to capture global structural information, such as specific graphic placements and non-uniform texture patterns across different UV islands. OmniFabric bridges this gap by shifting from local material extraction to global texture synthesis. We operate in the canonical UV space to disentangle base-color from illumination, synthesizing a holistic texture map that respects the global texture design in the reference image. While previous methods~\cite{zeng2024paint3d,bensadoun2024meta} also deploy a 2D refinement stage on the UV map, they only perform inpainting on occluded regions and fail to rectify distortion or remove baked-in artifacts, which are the domain specific challenges for garment texturing, as shown in Figure~\ref{fig:motivation}.

\label{sec:related}

%% file: sec/4_method.tex
\section{Method}

\begin{figure*}[t]
    \centering
    \includegraphics[width=0.98\textwidth]{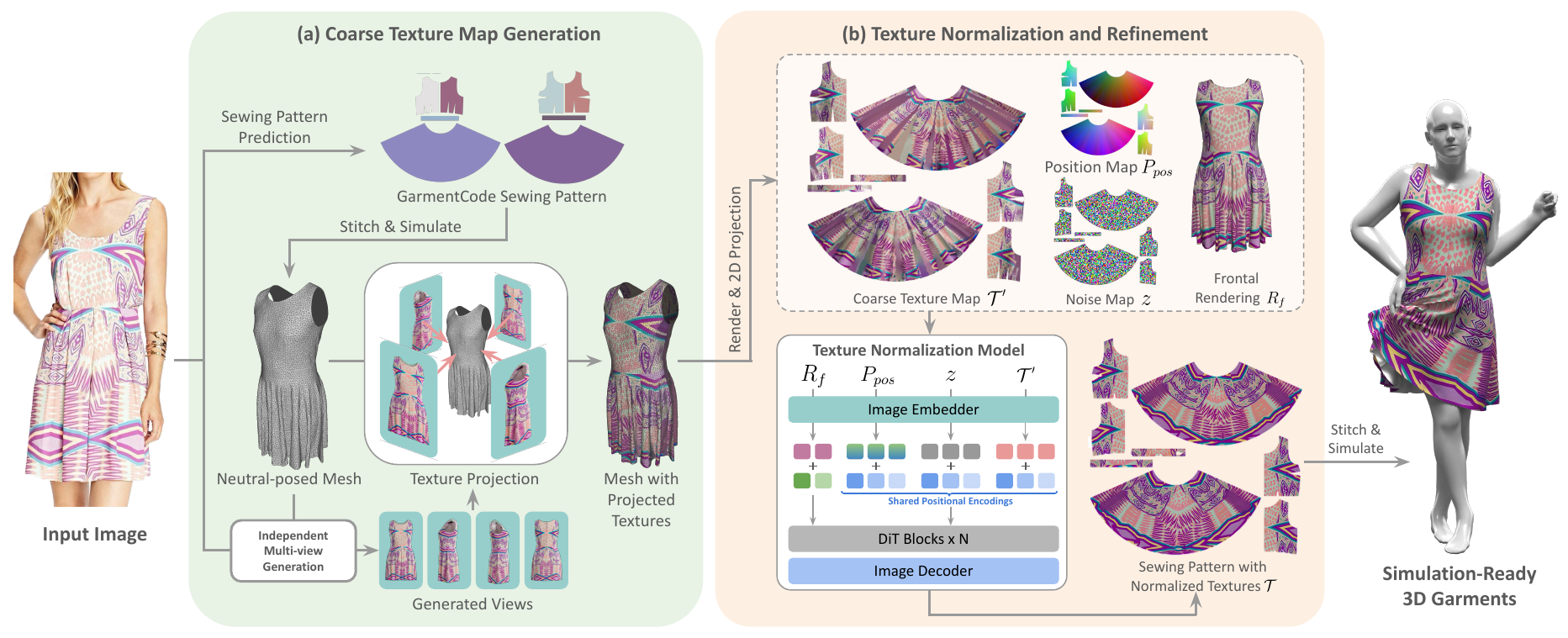}
    \vspace{-3mm}
    \caption{\small \textbf{Overview of OmniFabric.} Given a reference clothing image, OmniFabric synthesizes high-quality garment textures in sewing pattern UV space via a two-stage pipeline. \textbf{Stage 1:} We generate a coarse UV texture map by first predicting garment sewing patterns and reconstructing a canonical 3D garment. The input image texture is transferred onto the garment surface and expanded into multi-view using an off-the-shelf video generative model. These views are projected onto the canonical garment and reflected in UV space to produce a globally aligned coarse texture map.
  \textbf{Stage 2:} We refine the coarse texture with a DiT-based texture normalization network operating on UV patches. By conditioning on positional maps and spatially aligned tokens, the model removes baked-in artifacts and projection inconsistencies, producing clean and coherent sewing pattern textures that can be directly simulated into 3D garments.}
  \Description{A figure showing the overall pipeline of our method.}
    \label{fig:model_architecture}
    \vspace{-4mm}
\end{figure*}

Figure~\ref{fig:model_architecture} presents an overview of our approach. We aim to reconstruct 3D garments with coherent textures from real-world images. We define the task and objective in Section \ref{sec:problem_definition}. Next, we introduce a two-stage pipeline for generating holistic textures directly onto sewing patterns. This includes a coarse texture map initialization (Section \ref{sec:texture_projection}) and a UV space texture normalization (Section \ref{sec:texture_normalization}). To train the model, we develop an automated pipeline to synthesize a large scale dataset of textured sewing patterns (Section \ref{sec:dataset_creation}).

\subsection{Problem Definition}\label{sec:problem_definition}
Given a single reference image $I$ and an estimated 3D garment mesh $\mathcal{M_R}$ parameterized by sewing patterns $\mathcal{P}$ from an off-the-shelf sewing pattern prediction model, our goal is to generate a normalized texture map $\mathcal{T}$ directly within the sewing pattern space. 
Unlike previous methods~\cite{zhang2024fabricdiffusion} that focus on extracting local and repeatable material patches, we formulate our objective as a global texture synthesis task. We define $\mathcal{T} \in \mathbb{R}^{H \times W \times 3}$ as a representation of the normalized texture map of the garment, which is completely free of distortion, view-dependent illumination, cast shadows, and geometry-induced wrinkles. Following the paradigm of diffusion models, we formulate the synthesis of this holistic texture map as a conditional distribution mapping problem. We seek to learn a mapping function $\mathcal{G}$ such that:
\begin{equation} \label{mapping_function}
    \mathcal{T} \sim \mathcal{G}(\mathcal{I}, \mathcal{P}, \epsilon), \epsilon \sim \mathcal{N}(0, \mathbf{I})
\end{equation}

We note that although $\mathcal{T}$ is not intrinsic albedo, it's a close approximation and can be directly imported into cloth simulation engines, e.g., CLO~\cite{clo3d}, with externally specified material assumption for the remaining properties including roughness, metallic and normal map, enabling the synthesis of simulation-ready 3D garments.

\subsection{Coarse Texture Map Initialization}\label{sec:texture_projection}
To guide texture synthesis in the sewing pattern space, we first aim to obtain a coarse texture initialization that fully leverages the visual information from the reference image. This requires establishing pixel-wise correspondences between the reference image and the sewing pattern. While this mapping is challenging, we recast the task via a VLM-driven and pose aware texture alignment step. We further leverage the multi-view generative priors of large video models to provide a complete and globally coherent texture initialization.

\subsubsection{Pose aware texture alignment} Using Nano Banana Pro~\cite{nanobananapro}, we repose the reference image $I$ into $I'$ to perfectly align with the frontal silhouette of the rest-pose garment mesh $\mathcal{M}_R$. This alignment enables direct pixel projection of $I'$ onto $\mathcal{M}_R$ and its corresponding sewing patterns $\mathcal{P}$, producing the textured frontal render $R_f$. We observe that Nano Banana Pro successfully manipulates the input garment into a desired pose while preserving the original texture with high accuracy and fidelity.

\begin{figure*}[t]
    \centering
    \includegraphics[width=0.98\textwidth]{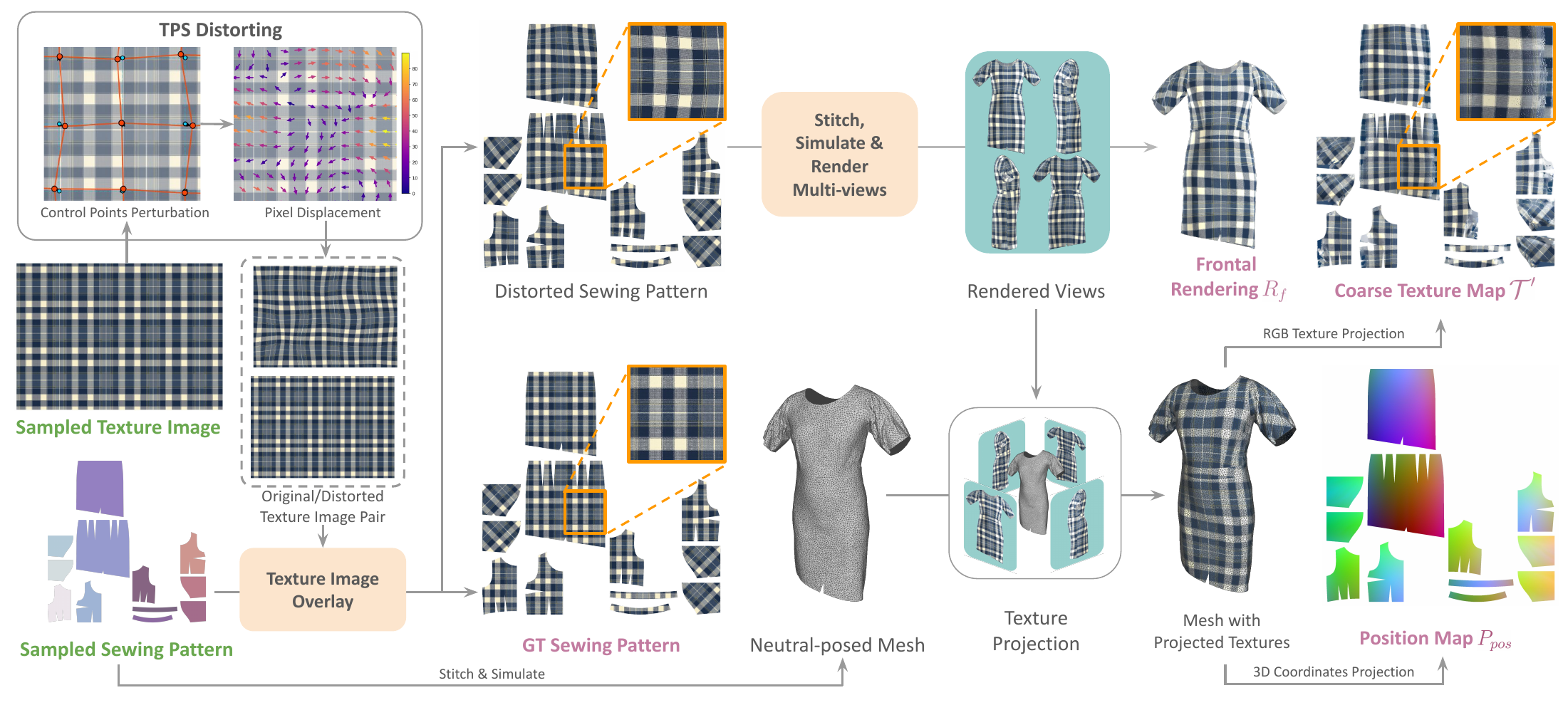}
    \vspace{-3mm}
    \caption{\textbf{Framework of automated data creation.} We first create a distorted texture image by applying a TPS warp to a sampled normalized texture. Both the normalized and distorted textures are then overlaid onto sampled sewing patterns. The textured sewing pattern can then be simulated into a training data point of rest-posed mesh $\mathcal{M_R}$, position map $P_{pos}$, coarse texture map $\mathcal{T}'$ and frontal view rendering $R_f$.
    }
    \Description{A figure showing our data curation pipeline.}
    \vspace{-1mm}
    \label{fig:data_pipeline}\vspace{-3mm}
\end{figure*}

\subsubsection{Consistent multi-view synthesis.} After the initial pose alignment, a significant portion of the UV space occluded from the frontal view remains untextured. To hallucinate these missing regions while encouraging strong spatial continuity, we leverage a large-scale video generation model for multi-view generation. By conditioning the model on the textured frontal render $R_f$, we synthesize a 360-degree spinning video of the garment, naturally exploiting the model's inherent temporal priors for structural consistency. Following recent projection-based texturing paradigms, we sample multi-view $V$, four orthogonal views (front, back, left, and right) from the generated video. These observations are then projected onto $\mathcal{M}_R$ and mapped to their corresponding UV coordinates on $\mathcal{P}$ to form the coarse texture map $\mathcal{T}'$.

\subsection{UV Space Texture Refinement}\label{sec:texture_normalization}

While $\mathcal{T'}$ preserves the coherent global texture of the garment, it contains baked-in artifacts like shadows, physical wrinkles and unpainted areas. Furthermore, spatial distortions in $\mathcal{T'}$ leave it far short of production-ready fabric quality, as shown in Figure~\ref{fig:motivation}. To address these issues, we introduce a texture normalization model designed to rectify $\mathcal{T'}$ into a normalized texture map $\mathcal{T}$.

\subsubsection{Training objective of multi-condition diffusion}
We formulate texture normalization as a holistic UV synthesis problem by developing a conditional distribution mapping network. Unlike FabricDiffusion~\cite{zhang2024fabricdiffusion}, which relies on a single condition, i.e., a local textile patch, for the normalization task, our framework incorporates a multimodal conditioning set $y$ to guide global texture normalization. We extend Eq.~\ref{mapping_function} and minimize the following:
\begin{equation}
    \mathcal{L} = \mathbb{E}_{\mathcal{E}(\mathcal{T}), y, \epsilon \sim \mathcal{N}(0,\mathbf{I}), t} \left[ \| \epsilon - \epsilon_\theta(z_t, t, y) \|^2 \right], y = \{ \mathcal{T}', R_f, P_{pos} \}
\end{equation}
where $z_t$ represents the noisy latent of the ground-truth texture at timestep $t$. The multi-modal conditioning set $y$ includes the coarsely-textured map $\mathcal{T'}$, textured frontal rendering $R_f$ and a 3D position map $P_{pos}$, created by projecting the vertices' 3D positions of $\mathcal{M_R}$ onto the 2D layout of $\mathcal{P}$. $P_{pos}$ helps the model learn spatial connectivity between separate UV panels, ensuring seamless textures across fragmented UV islands.

\subsubsection{Model architecture and training}
To process multimodal conditioning set $y$, we adopt a Diffusion Transformer (DiT) architecture following previous works~\cite{tan2025ominicontrol,tan2025ominicontrol2}, and leverage a unified token-processing strategy that treats all conditions as a unified sequence of tokens. We utilize a joint self-attention mechanism where the condition tokens (from $\mathcal{T}'$, $R_f$, and $P_{pos}$) and the noisy latent tokens from $z_t$ interact within the same transformer blocks. Because $\mathcal{T}'$, $P_{pos}$, and the target $\mathcal{T}$ all share the same 2D panel layout, we apply a dynamic positional encoding strategy: image patches at the same spatial coordinates across these maps are assigned identical positional encodings, while only $R_f$ retains its own coordinate system through distinct encodings. This minimal yet universal design allows the model to attend to the structural cues of $P_{pos}$ while simultaneously hallucinating missing textures in $\mathcal{T}'$. We use a pretrained weight of DiT, and fine-tune the model on our synthetic textured sewing pattern data with LoRA~\cite{hu2022lora}.

\section{Synthetic Training Data Creation}\label{sec:dataset_creation}

To train our texture normalization model, we develop a data engine (Figure~\ref{fig:data_pipeline}) to curate a large-scale dataset of textured sewing patterns. The core objective is to synthesize training pairs that exhibit the complex appearance of real-world garments while providing ground-truth normalized texture maps for supervised learning.

\subsection{Textured Sewing Pattern Creation}
A key bottleneck in garment dataset construction is the lack of complex and high-resolution textures; most available texture sources provide only simple, repetitive patterns. To overcome this, we introduce a pipeline to produce diverse, high-fidelity texture images by using fashion images from real-world datasets, e.g., DeepFashion~\cite{liuLQWTcvpr16DeepFashion}, as references. For each sample, we apply a Large Language Model (LLM) \cite{bai2023qwen} to generate a detailed description encompassing texture layout, semantic elements, and color palettes. These descriptions, combined with the reference image, are fed into a vision-language model \cite{nanobananapro} to synthesize textures with diverse designs. We then overlay the generated texture image onto sewing patterns sampled from GarmentCodeData. Note that the goal here is not to reconstruct the exact texture from the example image, but to generate plausible and realistic textures that provide sufficient diversity for training the texture synthesis model.

\subsection{Training Pairs Construction}
For each textured sewing pattern, we generate a training tuple $(\mathcal{T}', R_f, P_{pos}, \mathcal{T})$ consisting of a coarse texture map, a frontal rendering, a positional map, and the ground-truth texture. To explicitly train the network to rectify geometric distortions, we apply a random Thin Plate Spline (TPS) \cite{bookstein1989principal} distortion to the initial texture image. 
The original, undistorted texture forms the ground-truth map $\mathcal{T}$, while the TPS-distorted texture is overlaid onto the sewing pattern
  for physical simulation. Specifically, we perturb each control point $p_i$ of a regular grid fitted to the texture's aspect ratio, giving
  $\hat{p}_i = p_i + \delta_i$ with
  \begin{equation}
      \delta_i = \varepsilon_i \cdot \lambda \cdot \frac{D}{800}, \varepsilon_i \sim \mathcal{U}(-0.5, 0.5)
      \label{eq:tps_jitter}
  \end{equation}
  where $\lambda=25$ by default controls the overall distortion magnitude and $D$ is the resolution of a texture image. A
  TPS is then fit between the regular grid $\{p_i\}$ and its perturbed counterpart $\{\hat{p}_i\}$ and
  applied densely to warp the full-resolution texture. We stitch and drape the sewing pattern onto an A-pose SMPL body to yield a 3D garment mesh $\mathcal{M}_R$, from which we capture the frontal rendering $R_f$. The positional map $P_{pos}$ is then derived by projecting the $xyz$ coordinates of the mesh vertices onto the 2D panel layout $\mathcal{P}$. Finally, we synthesize the coarse texture map $\mathcal{T}'$ by projecting $R_f$ and a randomly sampled view $R_a$ onto a neutral-posed mesh and unwrapping them back into the 2D pattern space. This explicitly bakes physics-induced artifacts and deformations into the initialization, compelling the model to learn undistortion and normalized base-color recovery.

\label{sec:method}

%% file: sec/5_experiments.tex
\section{Experiments}
\label{sec:experiments}

In this section, we present a comprehensive evaluation of OmniFabric through quantitative and qualitative analyses. We first detail the experimental setup, including curated synthetic dataset and metrics used for evaluation. We then demonstrate the effectiveness of our framework by comparing against SOTA methods for 3D texturing. Finally, we conduct ablation studies to validate our core architectural designs. Additional implementation details, extended qualitative results, our method's adaptability to other sewing pattern prediction methods besides ChatGarment~\cite{bian2025chatgarment}, examples of our curated garment data, and discussion on potential future directions are provided in the supplementary material.

\subsection{Setup}

\subsubsection{Datasets}
To train and evaluate our model, we construct a large-scale synthetic dataset of textured sewing patterns, as described in Section~\ref{sec:dataset_creation}. We sample a total of 3K unique and diverse garment samples from GarmentCodeData~\cite{korosteleva2024garmentcodedata}. We then use the pipeline in Figure~\ref{fig:data_pipeline} to generate texture images, and print them onto sampled sewing patterns, finally constructing a dataset of 30K textured sewing patterns with diverse appearances and structural designs (10 texture variations per garment style on average). We simulate garments using NVIDIA Warp~\cite{Macklin_Warp_A_High-performance_2022} with a per-sample random seed, so repeated simulations of the same sewing pattern produce distinct $\mathcal{M}_R$ meshes and, consequently, different baked-in wrinkles in $\mathcal{T}'$. For lighting, we sample random subsets of white point lights to introduce varied illumination and baked-in shading across the dataset. We train our model with this synthetic dataset, and keep a held-out set for quantitative comparison with other baseline models with a test/train ratio of 0.1. Besides synthetic data, we evaluate all methods on in-the-wild images, which is the major focus of this work. These in-the-wild images are obtained from DeepFashion~\cite{liuLQWTcvpr16DeepFashion} dataset and recent clothing-design images collected from the web. Due to the lack of ground truth in-the-wild images, we evaluate mainly through qualitative comparisons.

\subsubsection{Metrics}
We evaluate fidelity and structural coherence of the generated textures with a suite of standard metrics. We use LPIPS~\cite{zhang2018unreasonable} and DISTS~\cite{ding2020image} to measure visual similarity and structural consistency. We also compare with SSIM~\cite{wang2004image} and MS-SSIM~\cite{wang2003multiscale}, metrics employed to assess pixel-level structural accuracy. CLIP-score (CLIP-s)~\cite{gal2022image} measures the semantic alignment between the final textured garment and the reference images.

\subsubsection{Baseline methods}
% To evaluate the effectiveness of OmniFabric, 
We compare against several SOTA methods in 3D texturing and garment-specific texture transfer. Among them, Hunyuan3D-2.0~\cite{zhao2025hunyuan3d} and Paint3D~\cite{zeng2024paint3d} use multi-view generation to provide additional prior beyond the given single-view observation. FabricDiffusion~\cite{zhang2024fabricdiffusion} is capable of creating fabric textures with no baked-in artifacts by normalizing local textile patterns with a diffusion model.

\subsubsection{Implementation details}

\begin{figure*}[!t]
  \centering
  \includegraphics[width=0.85\textwidth]{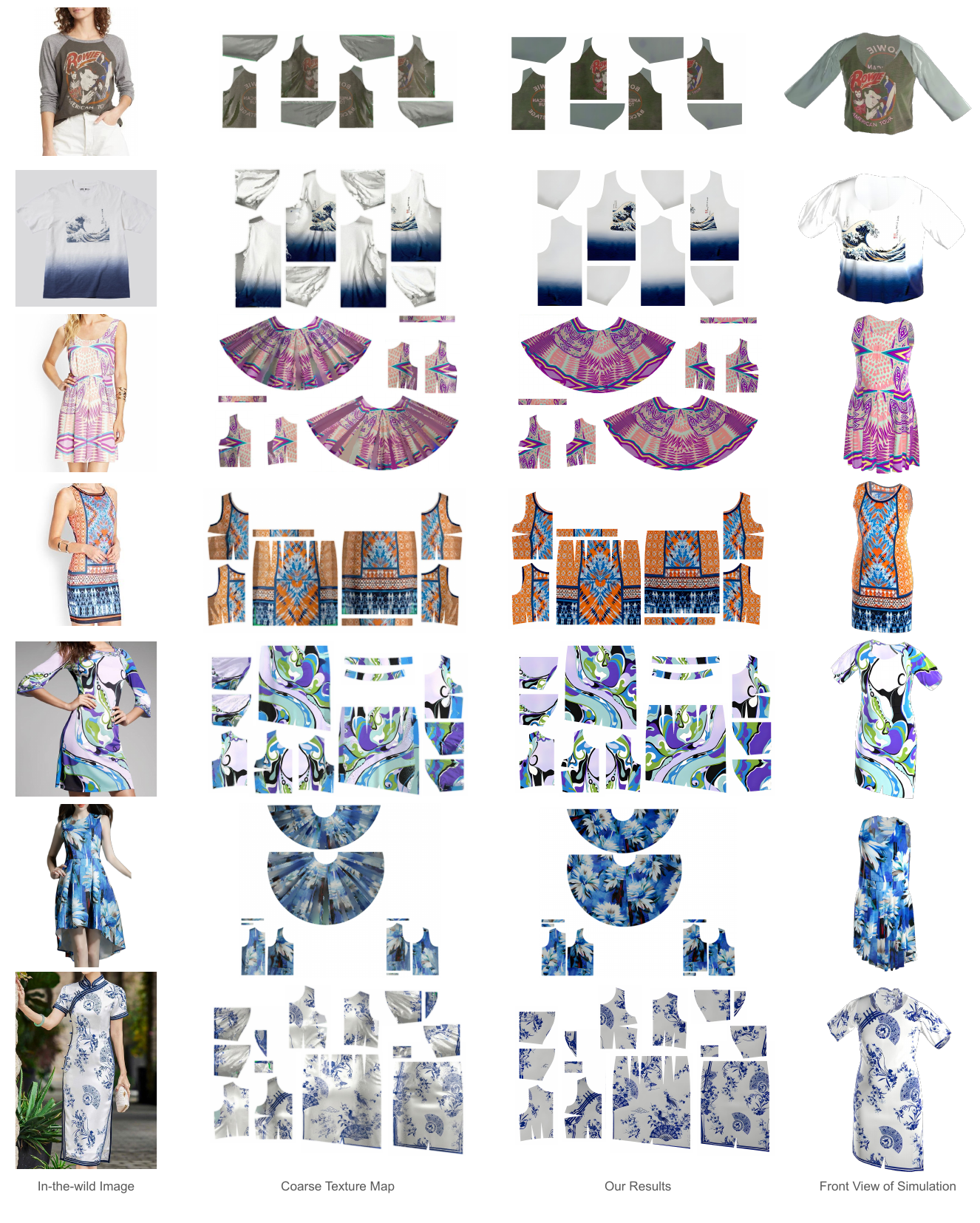}
  \vspace{-3mm}
  \caption{\textbf{Qualitative results of OmniFabric.} Our method synthesizes 3D garments with high appearance consistency by predicting sewing patterns of normalized textures. In particular, OmniFabric handles intrinsic geometric distortions and reconstructs diverse texture types, including logos and printed graphics (1st and 2nd rows), and complex high-frequency patterns with dense visual details. Please zoom in to check the details.
    }
    \Description{A figure showing the qualitative results.}
  \label{fig:qualitative_1}\vspace{-3mm}
\end{figure*}

\begin{figure*}[t]
  \centering
  \includegraphics[width=1\textwidth]{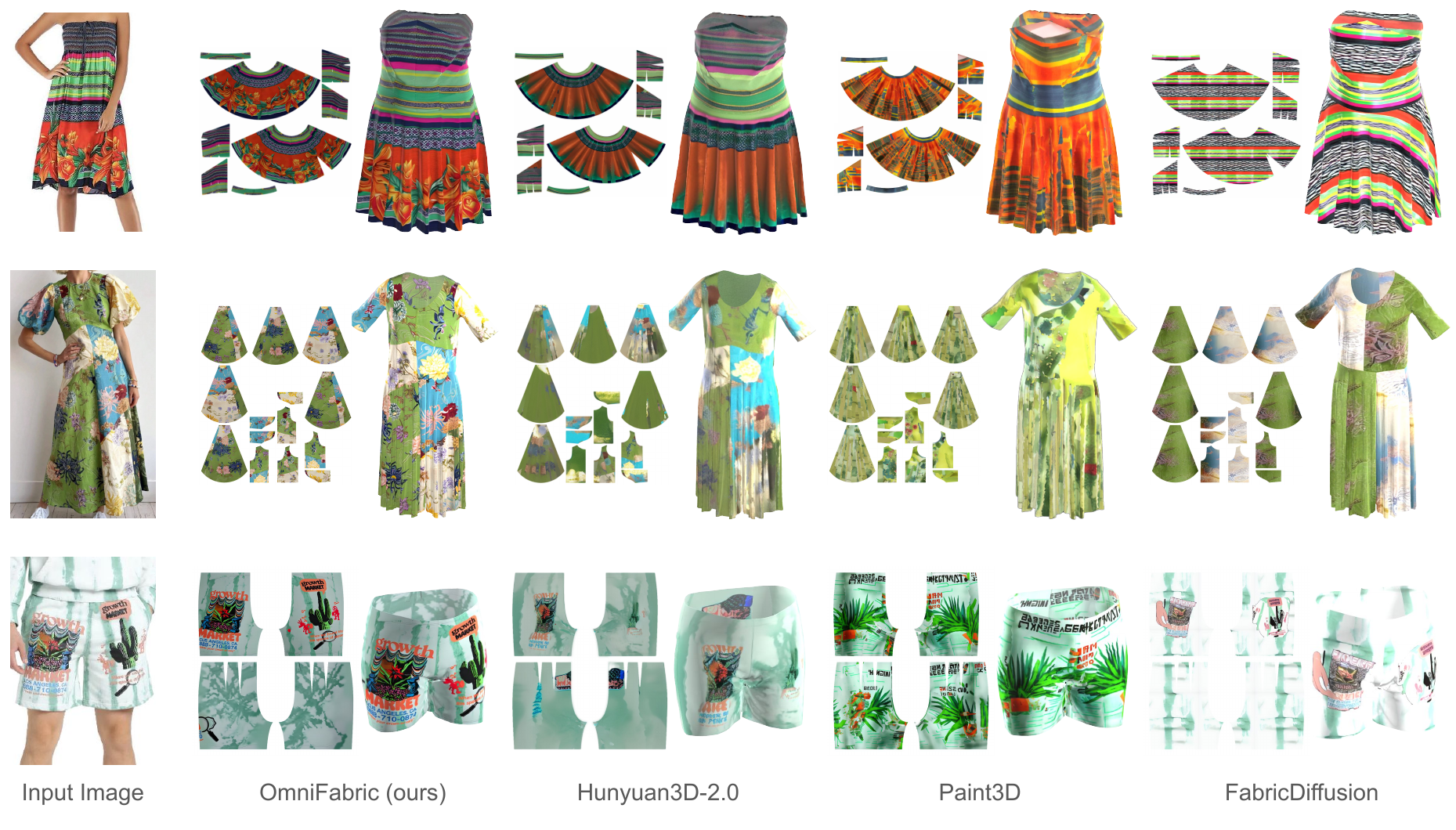}
  \vspace{-8mm}
  \caption{\textbf{Qualitative comparisons to state-of-the-art methods.} OmniFabric is capable of generating accurate asymmetric textures and logos, while existing methods fail to preserve the high-frequency details on input cases with complex patterns. 
    }
    \Description{A figure showing the qualitative comparisons among baselines.}
  \label{fig:qualitative_comparison}
  \vspace{-2mm}
\end{figure*}

Our texture normalization model is based on the Diffusion Transformer (DiT) architecture, and we follow the multi-conditioning strategy of OminiControl~\cite{tan2025ominicontrol} by treating tokens from all input conditions as a unified token sequence. The model is fine-tuned using LoRA on a single NVIDIA A6000 GPU with a batch size of 8, keeping the VAE and primary DiT backbone fixed to preserve generative stability. We resize and pad sewing pattern maps to a resolution of 1024$\times$1024 in a batch for training efficiency. During inference, we reverse the padding and resizing process, and deploy an image super-resolution model~\cite{wang2021real} to upscale the output. Since we are fine-tuning the model with LoRA, a higher training resolution can be easily achieved without GPU memory issues, making our framework scalable for training data of higher resolution. We fine-tune the model until convergence on our synthetic dataset. We utilize Veo 3~\cite{googledeepmind2025veo3web} as the multi-view video generation prior and Nano Banana Pro~\cite{nanobananapro} for the initial texture transfer to the frontal rendering $R_f$. We used ChatGarment~\cite{bian2025chatgarment} as the model for sewing pattern prediction throughout all experiments in the main paper. The average runtime for a single inference is roughly 5 minutes, depending on the current load on the Gemini model servers. The failure rate of Gemini models, e.g., generating unrelated textures on $R_f$ or multi-views, is lower than 2\%, estimated from randomly sampled generations. No additional filtering is necessary thanks to its stable performance. Additional details including prompts used for generation are provided in supplementary material.

\subsection{Qualitative Comparisons}
\subsubsection{Texture synthesis by OmniFabric from in-the-wild images}
We first show our results in Figure~\ref{fig:qualitative_1}. Given a reference image, OmniFabric first generates a coarse texture map, serving as a roughly textured sewing pattern. Then, the texture normalization model removes the distortions and baked-in artifacts. From the front views of simulated 3D garments provided in Figure~\ref{fig:qualitative_1}, we show that OmniFabric is capable of creating seamless textures across the 3D surface of garments. Note that while OmniFabric excels in textured garment synthesis from real-world images, we fine-tune our model only with synthetic data, which poses great scalability for model improvement.

\subsubsection{Simulation results}
We present simulations of synthesized 3D garments in Figure~\ref{fig:simulations} via CLO. The sewing patterns generated by OmniFabric provide the normalized RGB base-color as albedo. Other parameters not predicted by our model—roughness, metallic, reflection intensity, and auto-generated normal map—are left at CLO's default \texttt{Fabric\_Matte} preset values. The Environment/HDRI maps for relighting examples are also chosen from CLO's lighting presets. The simulated assets exhibit a convincing and photorealistic dynamic appearance, maintaining structural integrity and texture coherence even under extreme dynamics and varying illumination. 

\subsubsection{Comparison with state-of-the-art methods}
We compare with other methods in Figure~\ref{fig:qualitative_comparison} using in-the-wild images as well. For input with complex textures and elements, both Hunyuan3D and Paint3D fail to preserve the high-frequency details. Even though FabricDiffusion can create normalized textile pattern and texture the given mesh by tiling local textile patches, it is limited to local textile patches and fails to generate coherent global appearance. Conversely, OmniFabric creates normalized textures while preserving coherent details.

\subsection{Quantitative Comparisons}
\label{sec:quantitative_comparison}

\subsubsection{Synthetic dataset}
We show quantitative results where all baselines are evaluated with our curated synthetic dataset for fair comparison. For each test case, we provide every baseline with the identical frontal rendering $R_f$ and rest-posed mesh $\mathcal{M_R}$. Each method then performs its respective 3D texturing task on $\mathcal{M_R}$. To compute image-based metrics, the resulting 3D textures are projected back into the sewing pattern UV space to generate a comparable texture map $\mathcal{T}$ for each baseline. Table~\ref{tab:quant_1} shows OmniFabric excels in all metrics, particularly in preserving global structural coherence and eliminating projection artifacts.

\subsubsection{Real-world data}

\input{tables/quant_1}

\input{tables/user_study}

Besides comparing with other methods in Figure~\ref{fig:qualitative_comparison}, we conduct a user study to evaluate the performances of the methods due to the absence of real-world dataset of textured sewing patterns. As detailed in Table~\ref{tab:user_study}, the study involves 13 participants evaluating results from 10 real-world image inputs based on four criteria: overall quality ranking (1 to 4, lower is better), fidelity to the input appearance, back-view plausibility, and global texture coherence (1 to 5, higher is better). OmniFabric achieves the best results across all criteria, demonstrating that it is preferred on real images and more faithfully preserves coherent garment textures.

\subsection{Reproducibility via Open-Source Models}

\input{tables/open_source}

\input{tables/error_accumulation}

\input{tables/ablation_1}

While using Gemini models for pose transfer and multi-view generation (Section \ref{sec:texture_projection}) as default implementation of OmniFabric, we further construct a fully open-source alternative to facilitate reproducibility. To replace the closed-source models, we use FLUX.2~\cite{flux-2-2025} as image generation backbone and incorporate Ministral 3~\cite{liu2026ministral} for prompt up-sampling. We feed textured frontal view $R_f$ and silhouettes of four orthogonal views as conditions to generate the multi-view with this alternative. Table~\ref{tab:open_source} shows that open-source alternative achieves similar performance with the same experiment settings in Section \ref{sec:quantitative_comparison}.

\subsection{Error-Accumulation Analysis}

We analyze error accumulation across modules in our model architecture in this section by progressively replacing intermediate outputs, specifically the reposed reference image $I'$ and generated multi-view $V$, with ground-truth counterparts. As shown in Table~\ref{tab:error_accum}, settings (a) to (c) accumulate errors across stages including texture projection, pose alignment and multi-view generation, and (d) shows that our normalization model effectively fixes these aggregated artifacts on sewing patterns.

\subsection{Ablation Studies and Analyses}

\subsubsection{Importance of the canonical 3D and coarse texture map}
As shown in Table~\ref{tab:ablation}-(a), without the coarse texture map, which serves as the distillation of generative priors, the performance drops significantly. This shows that the projected textures do provide essential guidance for the texture generation. Figure~\ref{fig:ablation} shows that the model is randomly generating textures that ``look similar'' instead of pixel-aligned content without coarse texture map $\mathcal{T'}$.

\subsubsection{Effect of the frontal view rendering condition}
We observe that removing $R_f$ during training leads to inferior results (Table~\ref{tab:ablation}-(b)), suggesting that the frontal-view rendering provides important cues for the model. In particular, $R_f$ offers a holistic visual reference of the garment appearance, helping the model better understand global layout and consistency of the texture across sewing pattern pieces.

\subsubsection{Effect of position map}
As shown in Table~\ref{tab:ablation}-(c), without $P_{pos}$, the model not only shows suboptimal results but also textures and elements in wrong positions. As illustrated in Figure~\ref{fig:ablation}, a lemon is mistakenly generated in a blank region of the sewing pattern.

\subsubsection{Importance of geometric distortion in dataset.}
The right side of Figure~\ref{fig:ablation} shows the necessity of creating geometric distortion in the synthetic training data. Through training to remove geometric distortion, the model can not only generate normalized structured patterns on the sleeves, e.g., the checker pattern, but also produce clean appearances for other elements, e.g., texts and logos.

\subsection{Limitations}

\begin{figure}[!t]
  \centering
  \includegraphics[width=1.0\columnwidth]{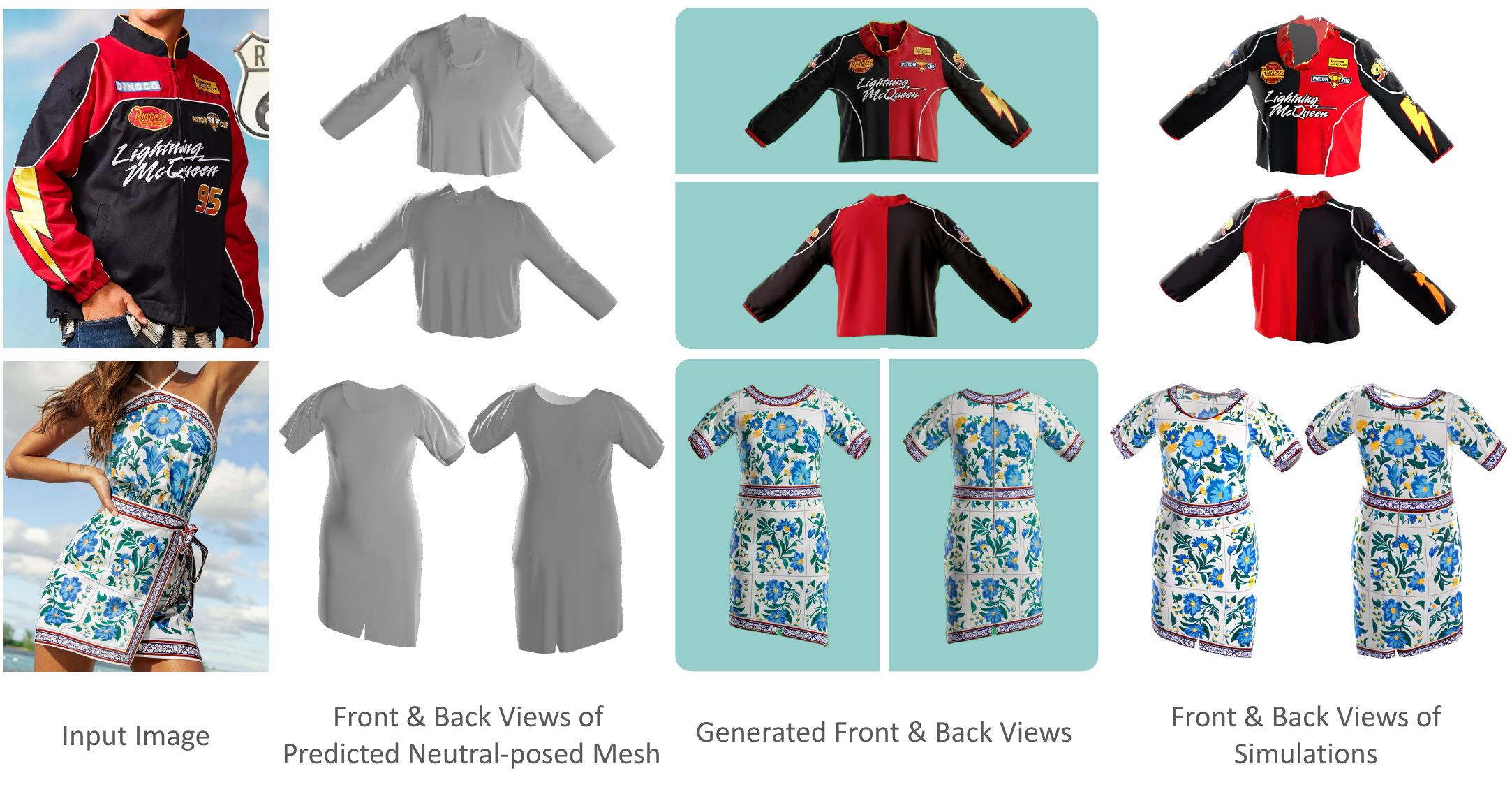}
  \vspace{-6mm}
  \caption{\textbf{Limitations.} OmniFabric may synthesize mismatched textures from inaccurately generated multi-views (1st row), or generate garments with incorrect geometry even though appearance is well-transferred (2nd row).} 
  \Description{A figure showing the failure cases of our method.}
  \label{fig:limitations}
  \vspace{-4mm}
\end{figure}

As shown in Figure~\ref{fig:limitations}, correctness of synthesized textures can be affected by inaccurately generated multi-views (1st row). OmniFabric is also unable to refine the mismatched geometry $\mathcal{M}_R$ predicted by ChatGarment~\cite{bian2025chatgarment} (2nd row); however, it does not change the role of our refinement model: given a sewing pattern, it normalizes the textures by reducing distortions, wrinkles, shadows, and projection artifacts, regardless of specific garment geometry. We put more discussion of limitations in the supplementary material.

%% file: tables/quant_1.tex
\begin{table}[!t]
  \centering
  \caption{\small \textbf{Quantitative comparisons.} We compare with baselines on our synthetic dataset. The results show that OmniFabric greatly surpasses all other methods in terms of both pixel and perceptual quality.}\vspace{-2mm}
  \small
  \resizebox{\columnwidth}{!}{
  \begin{tabular}{lccccc}
    \toprule
     &  LPIPS $\downarrow$ & SSIM $\uparrow$ & MS-SSIM $\uparrow$ & DISTS $\downarrow$ & CLIP-s $\uparrow$ \\
    \midrule
    FabricDiffusion~\cite{zhang2024fabricdiffusion} &  0.273 & 0.645 & 0.707 & 0.259 & 0.906 \\
    Paint3D~\cite{zeng2024paint3d} &  0.311 & 0.657 & 0.704 & 0.266 & 0.890 \\
    Hunyuan3D-2.0~\cite{zhao2025hunyuan3d} & 0.223 & 0.712 & 0.717 & 0.220 & 0.924 \\
    OmniFabric (ours)& \textbf{0.092} & \textbf{0.868} & \textbf{0.905} & \textbf{0.121} & \textbf{0.963} \\
    \bottomrule
  \end{tabular}
  }
  \label{tab:quant_1}
  \vspace{-2mm}
\end{table}

%% file: tables/user_study.tex
\begin{table}[h!]
  \centering
  \caption{\small \textbf{User study for real-world data.} We compare with baselines on real-world input images in 4 aspects: overall quality, fidelity, back-view plausibility and global coherence.}
  \vspace{-2mm}
  \small
  \resizebox{\columnwidth}{!}{
  \begin{tabular}{lcccc}
    \toprule
     &  Overall rank $\downarrow$ & Fidelity $\uparrow$ & Back-view $\uparrow$ & Global Coherence $\uparrow$  \\
    \midrule
    FabricDiffusion~\cite{zhang2024fabricdiffusion} &  3.40 & 1.75 & 2.69 & 3.15  \\
    Paint3D~\cite{zeng2024paint3d} &  3.41 & 1.86 & 2.81 & 3.03  \\
    Hunyuan3D-2.0~\cite{zhao2025hunyuan3d} & 2.10 & 3.43 & 3.41 & 3.50 \\
    OmniFabric (ours)& \textbf{1.09} & \textbf{4.44} & \textbf{4.35} & \textbf{4.41}  \\
    \bottomrule
  \end{tabular}
  }
  \label{tab:user_study}
  \vspace{-2mm}
\end{table}

%% file: tables/open_source.tex
\begin{table}[h!]
  \centering
  \caption{\small We construct a fully open-source alternative and it achieves similar performance to the default settings using closed-source models, showing the great reproducibility of our framework.}\vspace{-2mm}
  \small
  \resizebox{\columnwidth}{!}{
  \begin{tabular}{lcccc}
    \toprule
     &  LPIPS $\downarrow$ & SSIM $\uparrow$ & DISTS $\downarrow$ & CLIP-s $\uparrow$ \\
    \midrule
    OmniFabric w/ default settings & 0.092 & 0.868 & 0.121 & 0.963 \\
    OmniFabric w/ open-source models & 0.114 & 0.847 & 0.149 & 0.955 \\
    \bottomrule
  \end{tabular}
  }
  \label{tab:open_source}
  \vspace{-2mm}
\end{table}

%% file: tables/error_accumulation.tex
\begin{table}[h!]
  \centering
  \caption{\small We analyze error accumulation across certain modules by replacing intermediate outputs in our model pipeline with ground-truth (GT) counterparts. ``Pred.'' and ``GT'' mean the intermediate output is provided by Gemini models and GT respectively, and "SP Norm." refers to sewing pattern normalization. Results show that our texture normalization model effectively removes the accumulated error.}\vspace{-2mm}
  \small
  \resizebox{\columnwidth}{!}{
  \begin{tabular}{lcccc}
    \toprule
      \textbf{Replacement Setting} & LPIPS $\downarrow$ & SSIM $\uparrow$ & DISTS $\downarrow$ & CLIP-s $\uparrow$ \\
    \midrule
    \quad (a) GT $I'$ + GT $V$ & 0.105 & 0.849 & 0.159 & 0.944 \\
     \quad (b) Pred. $I'$ + GT $V$ & 0.121 & 0.837 & 0.176 & 0.935 \\
     \quad (c) Pred. $I'$ + Pred. $V$ & 0.144 & 0.804 & 0.166 & 0.938 \\
     \quad (d) Pred. $I'$ + Pred. $V$ + SP Norm. &  0.092 & 0.868 & 0.121 & 0.963 \\
    \bottomrule
  \end{tabular}
  }
  \label{tab:error_accum}
  \vspace{-2mm}
\end{table}

%% file: tables/ablation_1.tex
\begin{table}[h!]
  \centering
  \caption{\small \textbf{Ablation study} shows that our designs are essential for the texture normalization model to generate position-aligned details with high fidelity.}\vspace{-2mm}
  \label{tab:ablation_1}
  \small
  \resizebox{\columnwidth}{!}{
  \begin{tabular}{lcccc}
    \toprule
    \textbf{Setting} & LPIPS $\downarrow$ & SSIM $\uparrow$ & DISTS $\downarrow$ & CLIP-s $\uparrow$ \\
    \midrule
    OmniFabric (ours) & \textbf{0.092} & \textbf{0.868} & \textbf{0.121} & \textbf{0.963}  \\
    \quad (a) w/o coarse texture map $\mathcal{T'}$ & 0.407 & 0.601 & 0.306 & 0.887 \\
    \quad (b) w/o frontal view rendering $R_f$ & 0.097 & 0.850 & 0.129 & 0.962 \\
    \quad (c) w/o position map $P_{pos}$ & 0.098 & 0.848 & 0.130 & 0.961 \\
    \bottomrule
  \end{tabular}
  }
  \label{tab:ablation}
  \vspace{-3mm}
\end{table}

%% file: sec/6_conclusion.tex
\section{Conclusion and Discussion}

We introduce OmniFabric, a novel framework for synthesizing 3D garments with normalized and globally coherent textures. By leveraging large-scale generative priors and training a texture normalization model, we effectively disentangle intrinsic albedo from view-dependent artifacts including baked-in shadows and wrinkles. Our method bridges the gap between single-view observations and simulation-ready 3D assets, maintaining structural integrity across complex sewing patterns. For this study, we focus on the synthesis of sewing patterns with normalized textures within the GarmentCode framework. An immediate and promising extension would be to consider the joint prediction of complete PBR material maps, including roughness, metallic, and normal maps, alongside the albedo. We believe this direction will further enhance the photorealism of the reconstructed garments under diverse environmental lighting, providing even greater utility for immersive digital content creation.

\label{sec:conclusion}

%% file: sec/additional_qual.tex
\begin{figure*}[!t]
  \centering
  \includegraphics[width=0.96\textwidth]{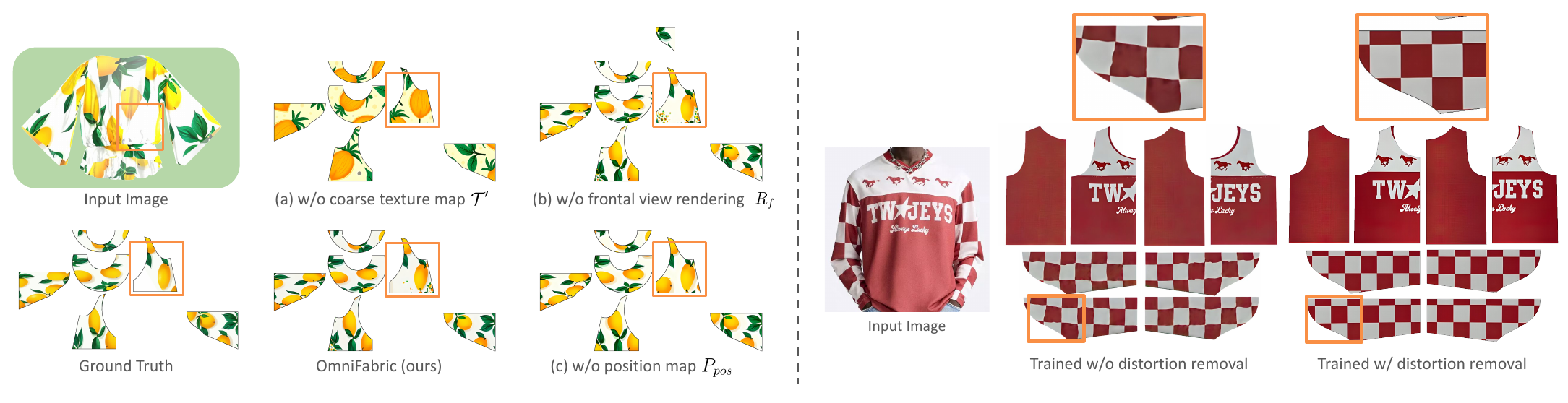}
  % \vspace{-3mm}
  \caption{\textbf{Qualitative analysis of key design mechanisms.} We demonstrate the effectiveness of the core designs of our texture normalization model. Without $\mathcal{T'}$, the model fails to generate consistent appearance but arbitrary textures with similar colors and elements. We also observe that $R_f$ and $P_{pos}$ are both essential conditions for the model to generate pixel-aligned content observed from the input image (left figure). We also show that our method can effectively rectify distortion when trained with distortion-augmented dataset (right figure). 
  % See Table~\ref{tab:ablation} for quantitative results.
  }
  \Description{A figure showing the ablation study of our method.}
  \label{fig:ablation}
\end{figure*}

\begin{figure*}[!t]
  \centering
  \includegraphics[width=0.9\textwidth]{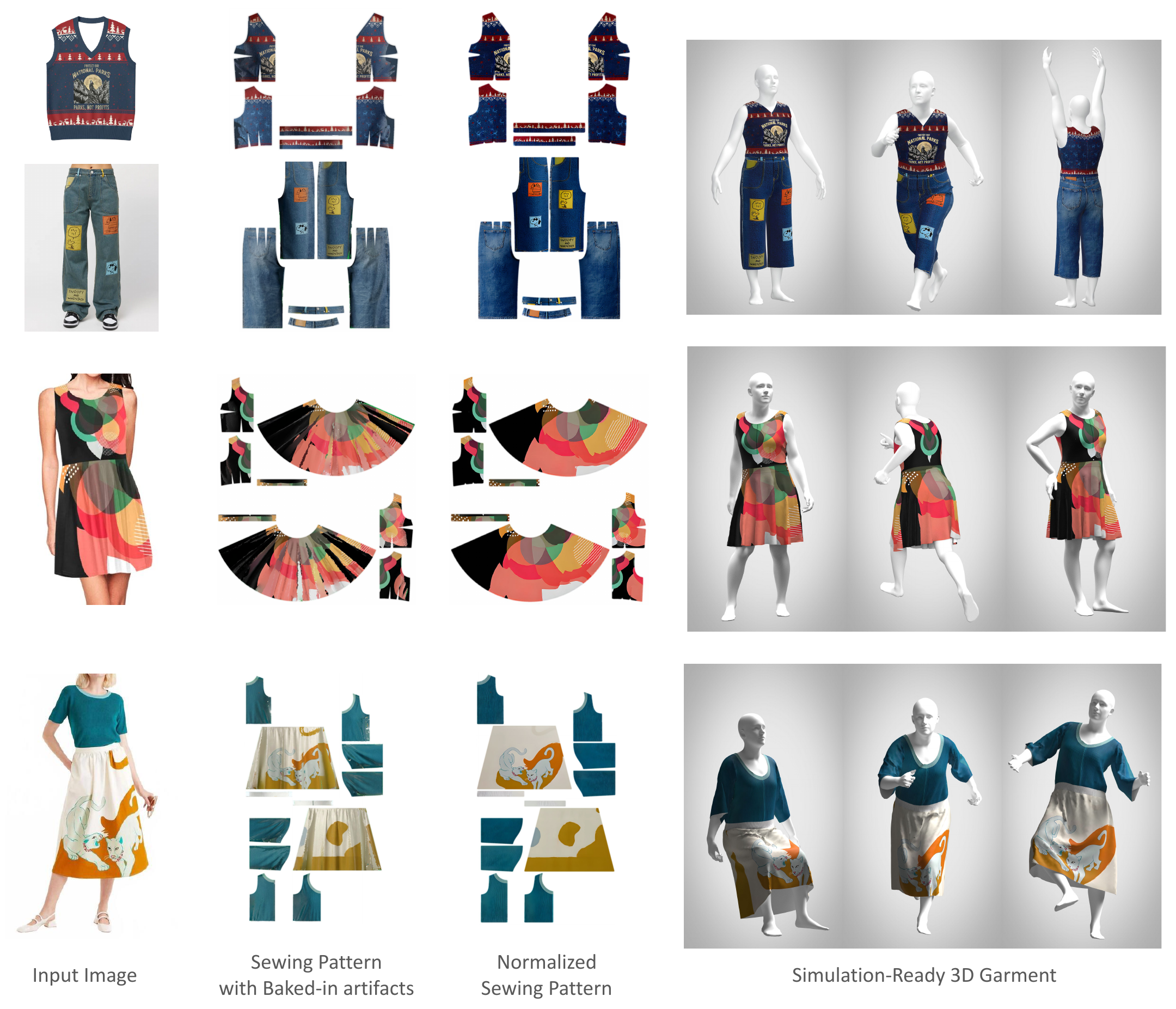}
  \vspace{-3mm}
  \caption{\textbf{Simulation results of our generated 3D garments.} 3D garments generated by our method can be directly simulated under varying human pose and lighting conditions and show convincing results.
    }
    \Description{A figure showing the simulation results of our synthesized 3D garment.}
  \label{fig:simulations}\vspace{-3mm}
\end{figure*}

%% file: sec/x_supp.tex
\section{Details of Dataset Construction}
\label{sec:details_of_dataset}

A primary contribution of this work is the automated data engine designed to generate high-fidelity textured sewing patterns. While high-detail manual texturing is possible, the process remains prohibitively time-consuming for large-scale applications. In contrast, our pipeline efficiently produces diverse garment styles and textures with a vast range of complexity, facilitating the construction of digital garment datasets at scale. This automated engine is a core contribution that addresses the scarcity of complex, high-resolution textures in existing garment research. We have included several representative examples in this supplement and will release the full dataset as well as the curation pipeline.

\begin{figure*}[h!]
  \centering
  \includegraphics[width=0.98\textwidth]{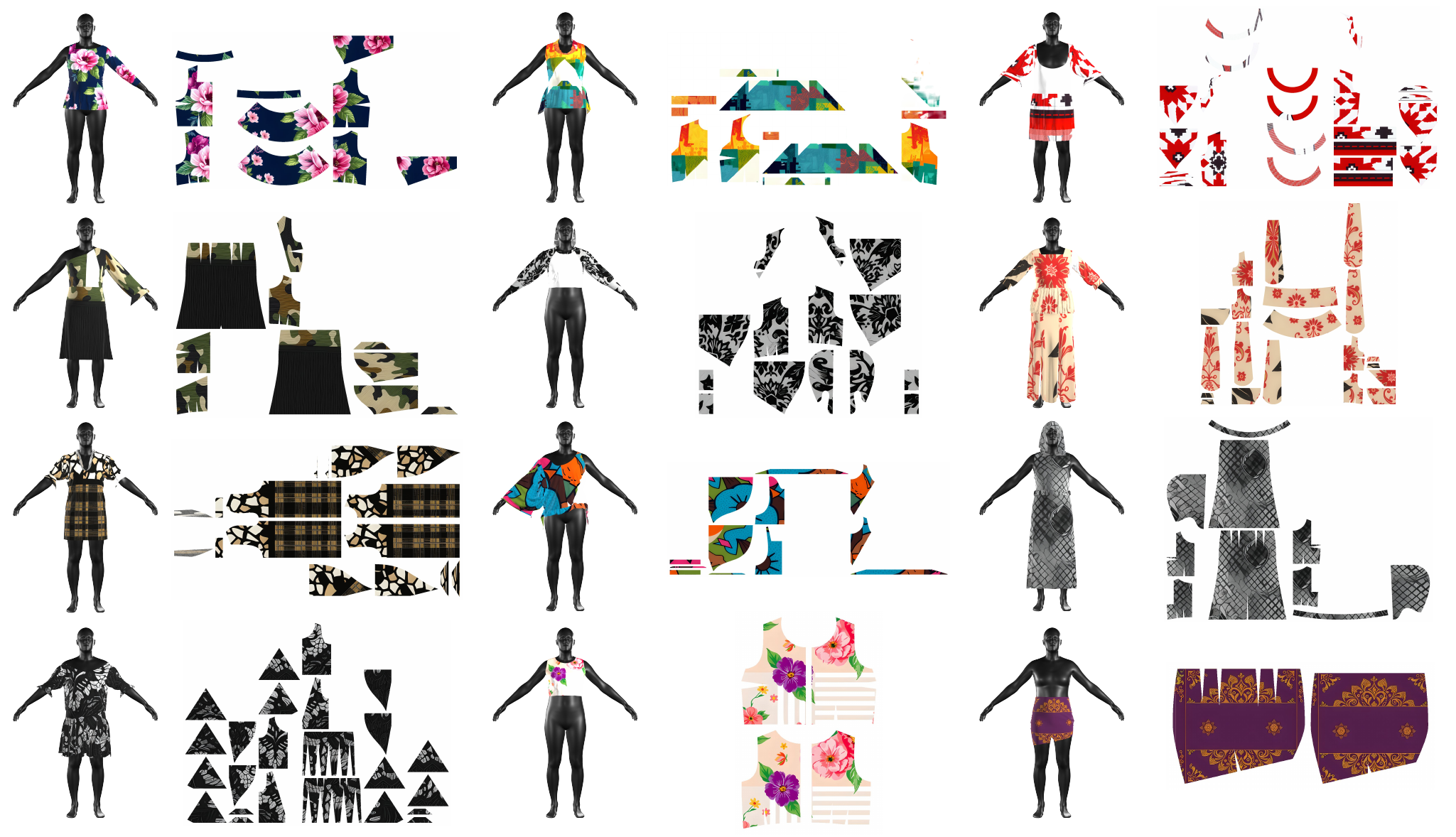}
  \vspace{-2mm}
  \caption{\textbf{Data examples of our synthetic dataset.} We show the synthesized textured sewing pattern and its simulated 3D garment for each datapoint.}
  \Description{A figure showing several examples in our curated dataset.}
  \label{fig:simulations_1}
  % \vspace{-4mm}
\end{figure*}

\subsection{Reordering Sewing Pattern for Texturing}

To align our generation process with real-world manufacturing, we introduce a reordering strategy for sewing pattern layouts. In professional garment construction, textures are often seamless across specific functional groups—such as a large frontal fabric panel—but exhibit discontinuities at structural seams, such as the transition from a sleeve to the chest. These discontinuities arise because 3D-stitched edges are often not geometrically matched in 2D space (e.g., a curved sleeve head joining a straight armhole), necessitating different fabric cuts.
Following these observations, our pipeline first reorders the 2D sewing pattern pieces by merging related panels, such as the front and back pieces of a single sleeve, into unified spatial groups. By overlaying textures onto this reordered 2D layout, we ensure seamless pattern continuity within each functional group while maintaining the realistic texture breaks required for production-ready 3D assets.

\section{Additional Implementation Details}
\label{sec:additional_implementation_details}

In this section, we provide further technical specifics regarding our texturing pipeline. As detailed in the main paper, we utilize a single-view garment observation to generate a 360-degree rotation video. From this sequence, we extract four keyframes corresponding to the orthogonal front, back, left, and right camera views for the multi-view texture projection process. Given that the video depicts a garment in a canonical A-pose, these four views are sufficient to capture the vast majority of surface details. The rest-posed mesh, $\mathcal{M}_{\mathcal{R}}$, is reconstructed by stitching and draping the predicted sewing patterns onto a A-posed SMPL human body model.

\subsection{Non-overlapping Texture Projection}

A key distinction between our projection method and state-of-the-art baselines, such as Hunyuan3D-2.0, lies in the source and application of multi-view data. While traditional methods project independent images generated by multi-view diffusion models, we project selected frames from a structurally consistent video prior using a specialized non-overlapping strategy. We observe that while multi-view diffusion models improve cross-view alignment, they often sacrifice high-frequency details to maintain that consistency. To preserve these details, our non-overlapping strategy begins by projecting textures from the front and back views—the two perspectives with minimal observational overlap—directly onto the mesh. Subsequently, textures from the side views are projected exclusively onto previously untextured regions, effectively performing a projection-based inpainting. Finally, our data-driven texture normalization model rectifies any minor artifacts or discontinuities at the seams of these projected textures. This approach allows OmniFabric to retain intricate, high-frequency surface details while ensuring global multi-view consistency.

\subsection{Prompt Configurations} 
In this section we specify the design of prompts for the Gemini models in our framework.
We use the prompt below for pose aware texture alignment, which generates a textured frontal view of rest-pose garment mesh $\mathcal{M}_R$:

\begin{quote}
\small
\texttt{The first apparel is my target apparel. Please generate textures on this apparel, so the textures are identical to the apparel in the reference image. Please don't change the style, size and pose of the target apparel.}
\end{quote}
For the multi-view generation with video model, we use the following prompt:
\begin{quote}
\small
\texttt{Create a continuous 360 degree rotation video of this apparel, making it rotate horizontally 360 degree, like a microwave.}
\end{quote}

\subsection{Training Details} We fine-tune all three baselines from pretrained weights, adapting the components described below. For Paint3D, we fine-tune its position encoder, as specified in its paper. For Hunyuan3D, we fine-tune its multi-view image generator. And for FabricDiffusion, we train its texture generator using paired textile data. All the models are fine-tuned upon pre-trained model weights.

\section{Additional Results and Analyses}
\label{sec:additional_results}

\begin{figure}[h!]
  \centering
  \includegraphics[width=1\columnwidth]{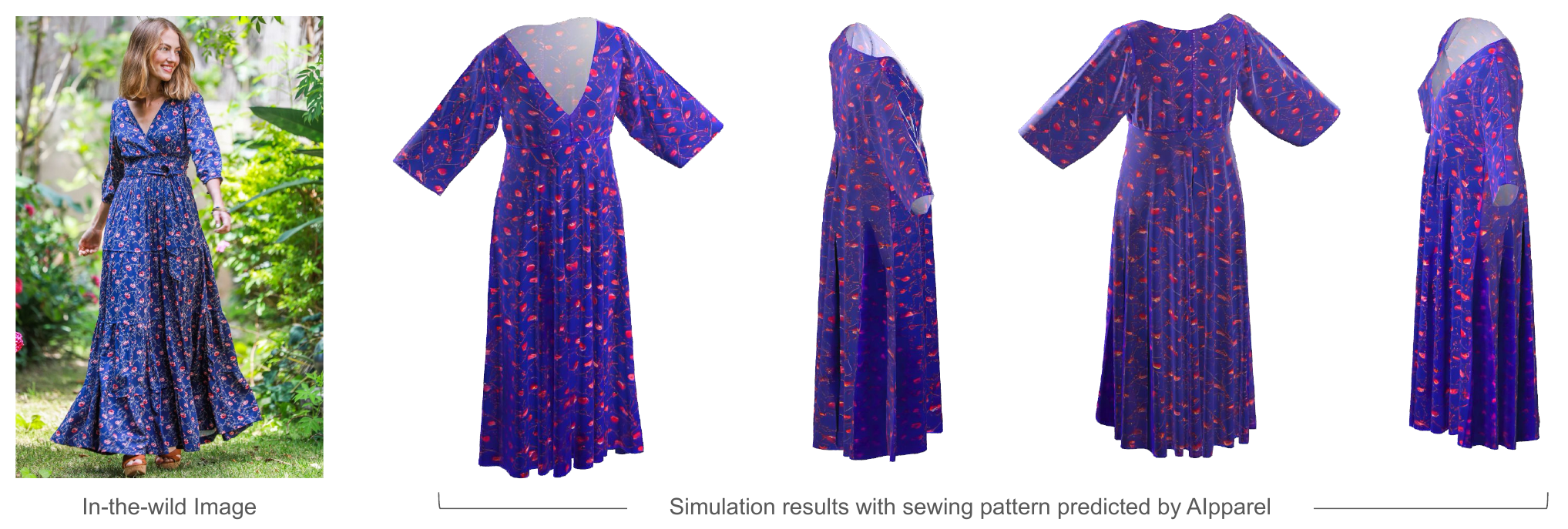}
  \vspace{-4mm}
  \caption{\textbf{Adaptability to other sewing pattern prediction method.} The simulation results show that OmniFabric can be seamlessly integrated with other methods of sewing pattern prediction. 
    }
    \Description{A figure showing that our method is compatible with other sewing pattern prediction methods.}
  \label{fig:aipparel}
  % \vspace{-4mm}
\end{figure}

\subsection{Robustness to Structural Variations in Sewing Patterns}
\label{sec:robustness}

\begin{figure*}[h!]
  \centering
  \includegraphics[width=0.98\textwidth]{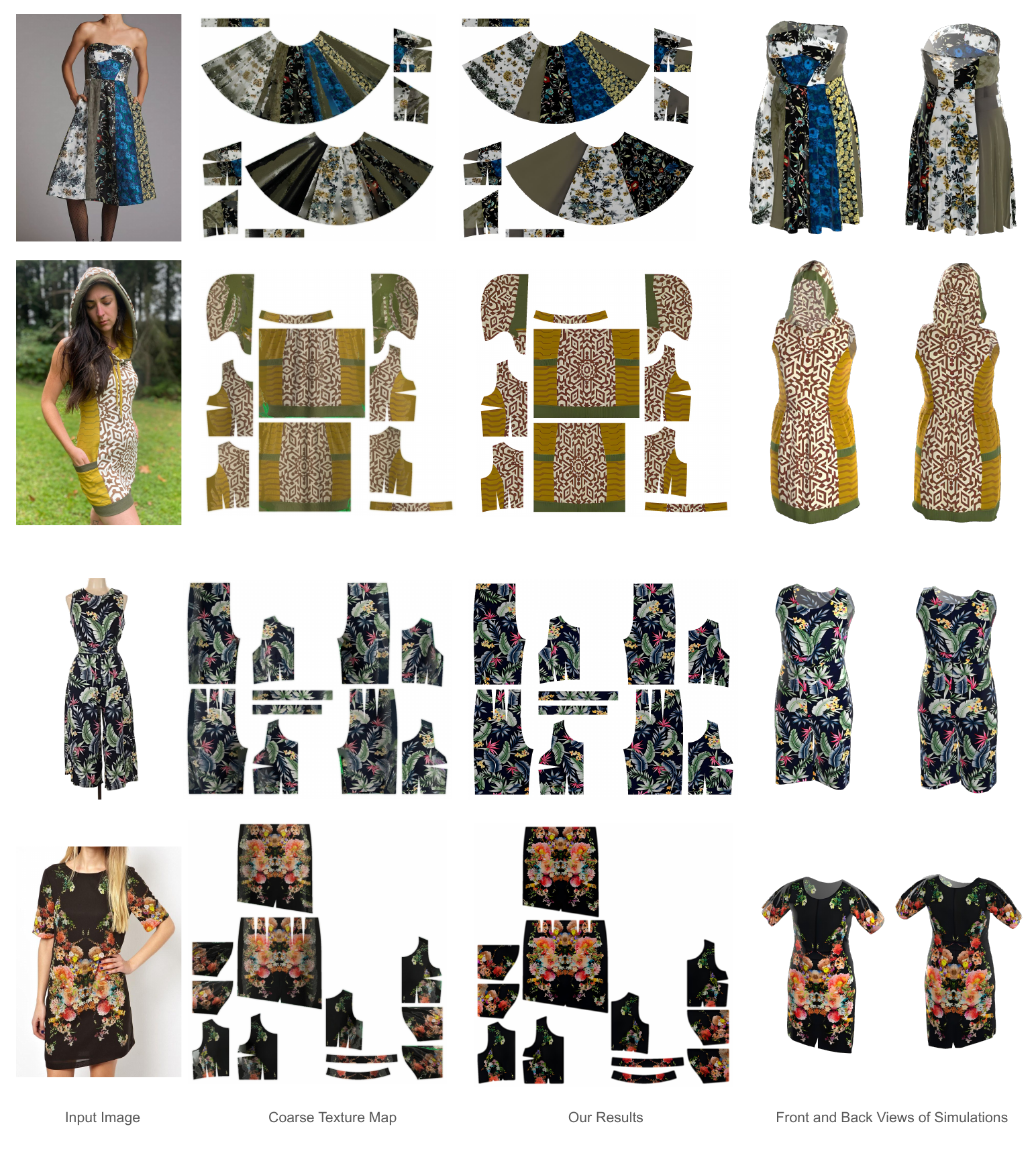}
  \vspace{-2mm}
  \caption{\textbf{Qualitative results.} OmniFabric effectively removes distortion artifacts and geometry-induced wrinkles, generating normalized textured sewing patterns that are simulation-ready.}
  \Description{A figure showing qualitative results of our method.}
  \label{fig:supp_qual_1}
  % \vspace{-4mm}
\end{figure*}

We present further qualitative evaluations in Figure~\ref{fig:supp_qual_1}, which demonstrate the robustness of our framework across a variety of garment styles. The results illustrate that our texture normalization model effectively rectifies the coarse texture map by eliminating projection-induced distortions and removing physics-based wrinkles inherent in the initial capture. While the geometry simulated from the predicted sewing patterns may occasionally deviate from the exact silhouette in the input image, we emphasize that such discrepancies arise from the limitations of the underlying sewing pattern prediction method rather than the texturing process. To mitigate this, our pipeline employs a texture transferring model that aligns the input image textures with the rendered silhouette of the predicted garment. This strategy successfully bridges the gap between the predicted geometry and the original observation, ensuring that the synthesized textures remain globally coherent and structurally aligned.

\subsection{Compatibility with Other Sewing Pattern Prediction Method}

While the main paper utilizes ChatGarment for sewing pattern prediction, our framework is designed to be agnostic to the specific reconstruction method employed. To demonstrate this flexibility, we evaluated our pipeline using sewing patterns predicted by Alpparel~\cite{nakayama2025aipparel}, a recent state-of-the-art approach based on Large Vision-Language Models. As illustrated in Figure~\ref{fig:aipparel}, OmniFabric consistently generates textured garments with aligned geometry and high-frequency details, confirming that our synthesis capability is not restricted to a single pattern reconstruction architecture.It should be noted, however, that the simulated garment geometry may occasionally exhibit minor mismatches relative to the original input image due to imperfect pattern prediction. As discussed in Section~\ref{sec:robustness}, our methodology specifically focuses on maintaining global texture consistency even when the underlying 3D mesh, which is simulated from predicted sewing pattern, is not perfectly aligned with the reference observation. This robustness ensures that OmniFabric can be seamlessly integrated with future advancements in sewing pattern prediction.

\section{Limitations and Future Work}
\label{sec:limitation}

While OmniFabric represents a significant advancement in garment texturing, several limitations remain that offer promising avenues for future research. First, although our model effectively removes transient lighting artifacts to produce a clean appearance, it does not yet explicitly disentangle albedo from shading in a strictly principled, physics-based manner, a distinction that should be noted to avoid overstating our current appearance decomposition capabilities. Furthermore, because the framework relies on generative priors, unseen or occluded regions are occasionally hallucinated; however, the system is not strictly limited to single-view observations and could incorporate multiple frames or video sequences in practice to reduce these hallucinations and improve reconstruction fidelity. Finally, certain materials with strong view-dependent effects, such as highly reflective leather or metallic fabrics, remain challenging to reconstruct faithfully, suggesting that future iterations should incorporate the joint prediction of complete PBR material maps—including roughness and normal maps—to achieve true photorealism across diverse environmental lighting conditions.